\documentclass[sigconf,screen,nonacm]{acmart}

\usepackage{multirow}
\usepackage{capt-of}
\usepackage{xspace}
\usepackage{arydshln}
\usepackage[table]{xcolor}
\usepackage{graphicx}
\usepackage{booktabs}

\usepackage{cleveref}
\usepackage{enumitem}
\usepackage{amsmath}
\usepackage{makecell}
\usepackage{array}
\usepackage{tabularx}
\usepackage{multirow}
\usepackage{makecell}
\usepackage{graphicx}
\usepackage{xcolor}
\usepackage{pifont}
\newcolumntype{L}[1]{>{\raggedright\arraybackslash}m{#1}}
\newcolumntype{C}[1]{>{\centering\arraybackslash}m{#1}}
\usepackage{tabularray}
\UseTblrLibrary{booktabs}

\definecolor{okgreen}{RGB}{0,140,70}
\definecolor{badred}{RGB}{200,35,35}

\renewcommand\footnotetextcopyrightpermission[1]{}
\title{Can AI Agents Really Complete RTL-to-GDS?\\
Lessons from Benchmarking Tool-Interactive EDA Workflows}

\author{%
  Jinyuan Deng$^{1}$,
  Zhengrui Chen$^{1,2}$,
  Xufeng Wei$^{1}$,
  Tianyu Xing$^{1}$,
  Chenyi Wen$^{1}$,
  Qi Sun$^{1}$,
  Cheng Zhuo$^{1,*}$ \\
  $^{1}$Zhejiang University, China \\
  $^{2}$ChipFlux, China \\
  $^{*}$Corresponding email: czhuo@zju.edu.cn
}

\begin{document}

\begin{abstract}
Large language model (LLM) agents are extending electronic design automation (EDA) beyond static RTL generation toward long-horizon, tool-interactive workflows. Yet it remains unclear whether general-purpose coding agents, even with domain-specific EDA skills, can reliably execute an end-to-end RTL-to-GDS flow encompassing synthesis, physical implementation, and engineering change order (ECO) optimization. We evaluate AI agents on a PicoRV32 RTL-to-GDS flow using commercial EDA tools under two timing targets. Their performance is assessed using end-to-end design score, stage completion, and Token ROI, a cost-efficiency metric relating design quality to runtime and cost. Comparing three agent architectures and four foundation models, we derive three practical lessons. First, domain-specific skills improve agents’ understanding of individual subtasks but do not ensure reliable completion of a long-horizon EDA flow. Second, agents that achieve similar design progress can still differ by up to 141× in Token ROI, revealing substantial differences in runtime and cost efficiency. Third, low-level tool-interface mismatches are a major source of physical design failures, particularly when Tcl commands depend on the tool version or execution mode. These results suggest that robust Agentic EDA requires not only stronger models but also structured tool interfaces, persistent design context, controlled execution, and process-level evaluation.
\end{abstract}


\maketitle

\section{Introduction}
Artificial intelligence is rapidly changing the role of automation in chip design. Earlier AI-for-EDA efforts largely focused on point solutions, such as quality-of-result (QoR) prediction, placement and routing optimization, or RTL generation. The emergence of large language models introduces a different possibility: an agent can reason about a design objective, call specialized tools, inspect reports and logs, revise its plan, and repeatedly act on the design until a verifiable artifact is produced. This transition shifts the system behavior from a feed-forward pattern of ``prompt--model--output'' toward a feedback loop of ``think/plan--act with tools--observe/reflect--re-plan.'' Recent community discussions therefore identify Agentic EDA as a promising path from isolated assistance to complex end-to-end automation~\cite{chen2026fluxeda,lu2025autoeda,team2026design,zhong2023llm4eda}.

RTL generation~\cite{thakur2024verigen,jin2025realbench} is an important entry point, but it is only the first step of a complete chip-design workflow. A usable design must proceed through logic synthesis, floorplanning, power planning, placement, clock-tree synthesis (CTS), routing, timing analysis, physical-aware ECO, and final artifact generation. Each stage produces a new design database, exposes a different action space, and introduces new constraints on timing, area, power, routability, and physical correctness. The flow is not a single programming problem. It is a long-horizon engineering process in which the validity of the next action depends on the current tool, loaded design, flow stage, analysis corner, and history of previous modifications.

This makes RTL-to-GDS a particularly revealing stress test for AI agents. In a conventional coding benchmark, a syntactically plausible command may be sufficient to obtain partial credit. In physical design, however, a single invalid command can terminate a session after a lengthy tool run, corrupt the expected design context, or invalidate downstream results. The difficulty is amplified by heterogeneous interfaces. Industrial EDA workflows commonly combine Tcl shells, Python scripts, Makefiles, vendor-specific commands, project conventions, and hand-maintained glue logic. Tool capabilities are often exposed through proprietary or version-dependent interfaces rather than a uniform action protocol. As a result, EDA flow integration remains fragmented and heavily dependent on expert application engineers.

The evaluation problem is equally important. Static RTL benchmarks measure whether generated code compiles, simulates, or passes hidden tests~\cite{liu2023verilogeval,lu2024rtllm,jin2025realbench,pinckney2025comprehensive}. These metrics are essential, but they do not reveal whether an agent can maintain a design context across many tool invocations, identify the stage at which a workflow failed, recover from an invalid action, or improve QoR without excessive trial and error. A final pass/fail label also hides large differences in engineering cost. Two agents may reach comparable final QoR, while one uses a small number of targeted actions and the other repeatedly regenerates scripts, reinitializes tools, and explores dominated configurations.

This paper asks a focused question: \textbf{Can current AI agents reliably complete an RTL-to-GDS design task under realistic and stringent design constraints?} We investigate this question through a representative case study: implementing PicoRV32~\cite{wolf2019picorv32} using commercial synthesis and physical-design tools under two timing targets with different levels of constraint tightness. The case study compares a general-purpose coding agent, the same agent augmented with EDA skills, and FluxEDA, an execution architecture designed to organize heterogeneous EDA capabilities into an executable, verifiable, and recoverable workflow. Our evaluation considers not only final design quality, but also stage completion, runtime, token cost, Token ROI, and physical-design failure patterns.

{Our contributions can be summarized as follows:
\begin{itemize}
\item We formulate RTL-to-GDS implementation under different constraints as a tool-interactive agent evaluation problem. The task requires agents to maintain design state, respond to tool feedback, and produce physically valid results across multiple design stages.
\item We develop a process-aware evaluation methodology that jointly measures stage completion, design quality, runtime, token cost, Token ROI, and failure patterns. This evaluation distinguishes inexpensive early failures from efficient and verified design progress.
\item We compare multiple agent architectures and foundation models, and show that reliable RTL-to-GDS completion depends on the complete agent system rather than the foundation model alone. Domain-specific skills provide inconsistent gains, while structured tool interfaces and persistent design context substantially improve execution stability and reduce operational failures.
\end{itemize}}

\section{Why RTL-to-GDS Is a Hard Agent Environment}
\subsection{Long-Horizon, Multi-Stage Execution}
The traditional RTL-to-GDS flow is a multi-stage engineering process. Logic synthesis transforms RTL and constraints into a gate-level implementation; floorplanning defines geometry, utilization, and major physical structures; placement and CTS determine cell locations and clock distribution; routing creates detailed interconnect; static timing analysis and ECO iteratively repair setup and hold violations; and final checks produce the netlist, layout, timing reports, and other deliverables. A failure late in the flow may invalidate hours of upstream work.

For an agent, the challenge is not simply to know the nominal sequence of steps. It must recognize which steps are already complete, which reports are current, which actions have side effects, and whether a modification improves the global objective or merely shifts a violation to another corner. For example, a timing fix that improves setup slack may degrade hold slack or increase area. A denser floorplan may reduce wirelength but create routing congestion. The agent must therefore account for delayed feedback and competing objectives while maintaining a consistent design state throughout the flow.

\subsection{Fragmented Tool Interfaces and Hidden Operational State}
Industrial EDA orchestration faces a long-standing integration bottleneck. Tool entry points are fragmented across proprietary Tcl, Python, Bash, and workflow scripts. Internal capabilities are frequently encapsulated behind tool-specific protocols, and external automation relies on project-local wrappers and manual CAD effort. This fragmentation creates two related problems for LLM agents.

First, the valid action space is narrower than the apparent programming language. A command can be syntactically well formed yet invalid for the active tool version, execution mode, database state, or multi-mode multi-corner (MMMC) initialization. Second, the operational state required to interpret a command is often implicit. The agent must know which design is loaded, whether clocks have been propagated, which constraints and parasitics are active, and which reports or artifacts correspond to the current database. Losing this context turns a continuous engineering process into disconnected one-shot scripts.

A robust agent execution layer must therefore provide controlled, stateful access to EDA tools rather than unrestricted shell access. It should expose tool capabilities through validated interfaces, maintain persistent tool sessions, return structured observations, preserve the association between reports and design states, and isolate each run from unrelated workspaces. These requirements motivate the execution architecture introduced in the following section.

\subsection{From Final Correctness to Process-Level Efficiency}
Agentic workflows substantially increase interaction volume. Compared with a one-shot generation setting, the agent must repeatedly inspect logs, request reports, propose actions, verify outcomes, and sometimes rollback. The cost is not only token consumption. Each unnecessary action can invoke an expensive EDA run, consume a license, and delay the overall design cycle. A benchmark that reports only the final QoR may therefore reward brute-force or unstable behavior.

We use Token ROI to summarize how effectively an agent converts model and tool investment into EDA progress. At a high level,
\begin{equation}
\mathrm{TokenROI}=\frac{\text{normalized EDA return}}
{\text{token investment}\times\text{execution-time investment}}.
\label{eq:roi}
\end{equation}
The numerator represents task-specific design quality and completion, while the denominator penalizes excessive model interaction and long tool runtime. We further use normalized Token ROI, which rescales token cost and execution time to comparable ranges for more consistent comparison across runs. A meaningful RTL-to-GDS agent evaluation should therefore consider
not only final design quality, but also stage completion, execution
cost, and failure behavior. Together, these process-level signals
distinguish efficient, verified progress from low-cost early failures
or expensive trial-and-error execution.

\begin{figure*}[ht]
    \centering
    \includegraphics[width=1.0\linewidth]{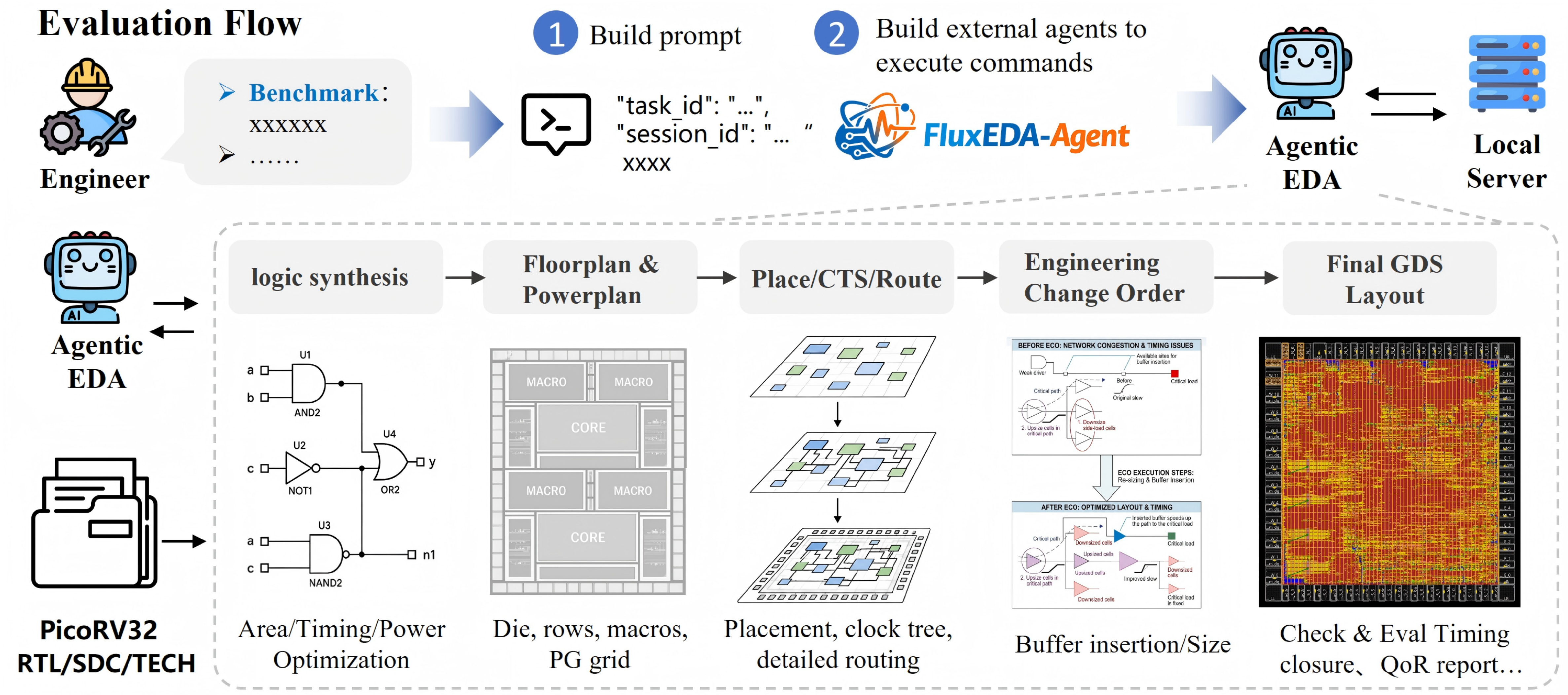}
    \caption{PicoRV32 RTL-to-GDS evaluation flow. The agent must advance a single design through synthesis, physical implementation, physical-aware ECO, and final deliverable generation while receiving tool feedback after each stage.}
    \label{Fig:oval_all_rtl_to_gds} 
\vspace{-0.2cm}
\end{figure*}

\subsection{Relation to Existing EDA Benchmarks}
Existing RTL benchmarks have driven rapid progress by providing reproducible specifications, simulators, and hidden tests. VerilogEval and RTLLM focus on functional RTL generation, while newer datasets such as RealBench and CVDP increase realism, coverage, and difficulty~\cite{liu2023verilogeval,lu2024rtllm,jin2025realbench,pinckney2025comprehensive}. These benchmarks answer an essential question: can a model translate a specification into functionally correct hardware description code?

RTL-to-GDS asks a complementary question. Even when the RTL is fixed and correct, can an agent navigate a heterogeneous implementation flow, adapt to tool feedback, and produce a physically meaningful result? This setting introduces delayed rewards, expensive evaluation, large combinatorial action spaces, and cross-stage dependencies. The same final QoR may be reached through very different trajectories, and many failures arise from the interaction boundary rather than the design algorithm itself. For this reason, RTL-to-GDS should be viewed not as a replacement for RTL generation benchmarks, but as a workflow-level extension that evaluates agent behavior under realistic engineering constraints.

A complete Agentic EDA benchmark will eventually need both. Code-grounded tasks test generation and repair against compilation or simulation feedback. Flow-grounded tasks test multi-stage orchestration and deliverable production. Report-grounded tasks test whether an agent can interpret timing, DRC, or power reports and perform targeted optimization. The present invited paper isolates the flow-grounded case so that the operational challenges and lessons can be examined in depth.

\section{Case Study and Evaluation Methodology}
\subsection{PicoRV32 RTL-to-GDS Task}
We select PicoRV32 as the target design  because it provides a more realistic multi-stage implementation workload than small arithmetic benchmarks, while remaining tractable for repeated commercial-tool experiments. The task begins with RTL, timing constraints, and technology configuration, and requires the agent to complete logic synthesis, floorplanning and power planning, placement, CTS, routing, physical-aware ECO, and final report generation. The expected outputs include a gate-level netlist, placed-and-routed layout data, timing and QoR reports, and the final GDS-related deliverables.

Two clock targets are used to vary task difficulty. The 350-MHz setting is a relatively loose target that tests whether an agent can complete the entire flow and produce valid outputs. The 700-MHz target places greater pressure on timing closure and makes the flow more sensitive to implementation and ECO decisions. The pair exposes whether an approach merely follows a nominal recipe or can continue to operate when the design objective becomes demanding.

\begin{table}[t]
\centering
\caption{PicoRV32 RTL-to-GDS case-study setup.}
\label{tab:setup}
\small
\setlength{\tabcolsep}{4pt}
\begin{tabular}{@{}p{0.29\columnwidth}p{0.61\columnwidth}@{}}
\toprule
\textbf{Item} & \textbf{Setting} \\
\midrule
Design & PicoRV32 RISC-V core \\
Technology & Commercial 55-nm technology setup \\
Clock targets & 350 MHz (loose), 700 MHz (tight) \\
Flow stages & Synthesis, floorplan/power, placement, CTS, routing, ECO, final reports \\
Architectures & Claude Code, Claude Code + EDA Skills, FluxEDA \\
Models & DeepSeek-V4, GLM-5.2, Claude Sonnet 5, Kimi K3 \\
Measurements & Score, Token ROI, cost, runtime, stage completion, error category \\
\bottomrule
\end{tabular}
\end{table}

\subsection{Agent Architectures}
\textbf{Claude Code(CC)} ~\cite{anthropic_claude_code_docs} represents a general-purpose command-line coding agent. It can inspect files, generate scripts, and interact with the tool environment through shell-oriented operations. This baseline tests how far a capable general agent can progress without an EDA-specific execution substrate.

\noindent\textbf{Claude Code + EDA Skills} ~\cite{ickylin_skills_2026} augments the same interaction pattern with domain-specific procedural knowledge. The skills provide guidance on flow steps, commands, reports, and common debugging procedures. This condition tests whether adding EDA knowledge alone is sufficient to stabilize the workflow.

\noindent\textbf{FluxEDA} ~\cite{chen2026fluxeda} separates agent reasoning from tool execution. The agent discovers registered EDA capabilities, invokes structured actions through a gateway, receives normalized results, and reuses persistent tool sessions and design context. The architecture is not another EDA engine; it is an execution harness that abstracts heterogeneous engines into a unified API, manages lifecycle and state, and constrains invalid or unsafe operations.

Each architecture is paired with DeepSeek-V4 ~\cite{xu2026deepseek}, GLM-5.2 ~\cite{glm52}, Claude Sonnet 5 ~\cite{anthropic_sonnet5} and Kimi K3 ~\cite{kimi_k3}. Holding the task and tool environment fixed while varying the architecture and foundation model helps distinguish model capability from execution support.

\subsection{Scoring Method}

Each run receives an end-to-end score on a 0--100 scale:
\begin{equation}
S=0.2S_{\mathrm{stage}}+0.8S_{\mathrm{result}}.
\label{eq:end-to-end-score}
\end{equation}
{The stage score measures sequential progress through logic synthesis,
physical implementation, and ECO. Logic synthesis is evaluated using
netlist-generation and clock-gating checks. Physical implementation
includes I/O placement, standard-cell placement, clock-tree synthesis,
and routing, while ECO is evaluated as a binary pass/fail stage. The
stages are gated sequentially: physical-implementation checks contribute
only after logic synthesis is complete, and the ECO stage contributes
only after both preceding stages have been fully passed. The stage score
is the average of the three resulting gate scores.
The result score is enabled only when all three stages are complete. Let}
\begin{equation}
\mathcal{M}=\{A,P,W_s,T_s,W_h,T_h\},
\end{equation}
where \(A\) and \(P\) denote area and power, respectively, and
\(W_s\), \(T_s\), \(W_h\), and \(T_h\) denote setup WNS, setup TNS,
hold WNS, and hold TNS, respectively. The result score is computed as
\begin{equation}
S_{\mathrm{result}}=
\begin{cases}
\displaystyle \frac{1}{6}\sum_{m\in\mathcal{M}}S_m,
& \text{if all three stages are complete},\\[6pt]
0, & \text{otherwise},
\end{cases}
\label{eq:result-score}
\end{equation}
where \(S_m\) denotes the normalized score of metric \(m\). Area and
power are smaller-is-better metrics and are normalized against
frequency-specific reference values. Timing metrics are
larger-is-better: nonnegative values receive full credit, while
negative values are scaled according to predefined lower bounds.
This gating rule prevents partial outputs or isolated reports from
being treated as valid final designs.

Token ROI complements the end-to-end score by relating normalized
design return to the token cost and execution time required to achieve
it, as defined in Eq.~\eqref{eq:roi}. For normalized Token ROI, token
cost and execution time are also rescaled before the efficiency score
is computed, reducing the effect of differences in scale between the
two cost components.

\begin{figure}[t]
\centering
\includegraphics[width=0.95\linewidth]{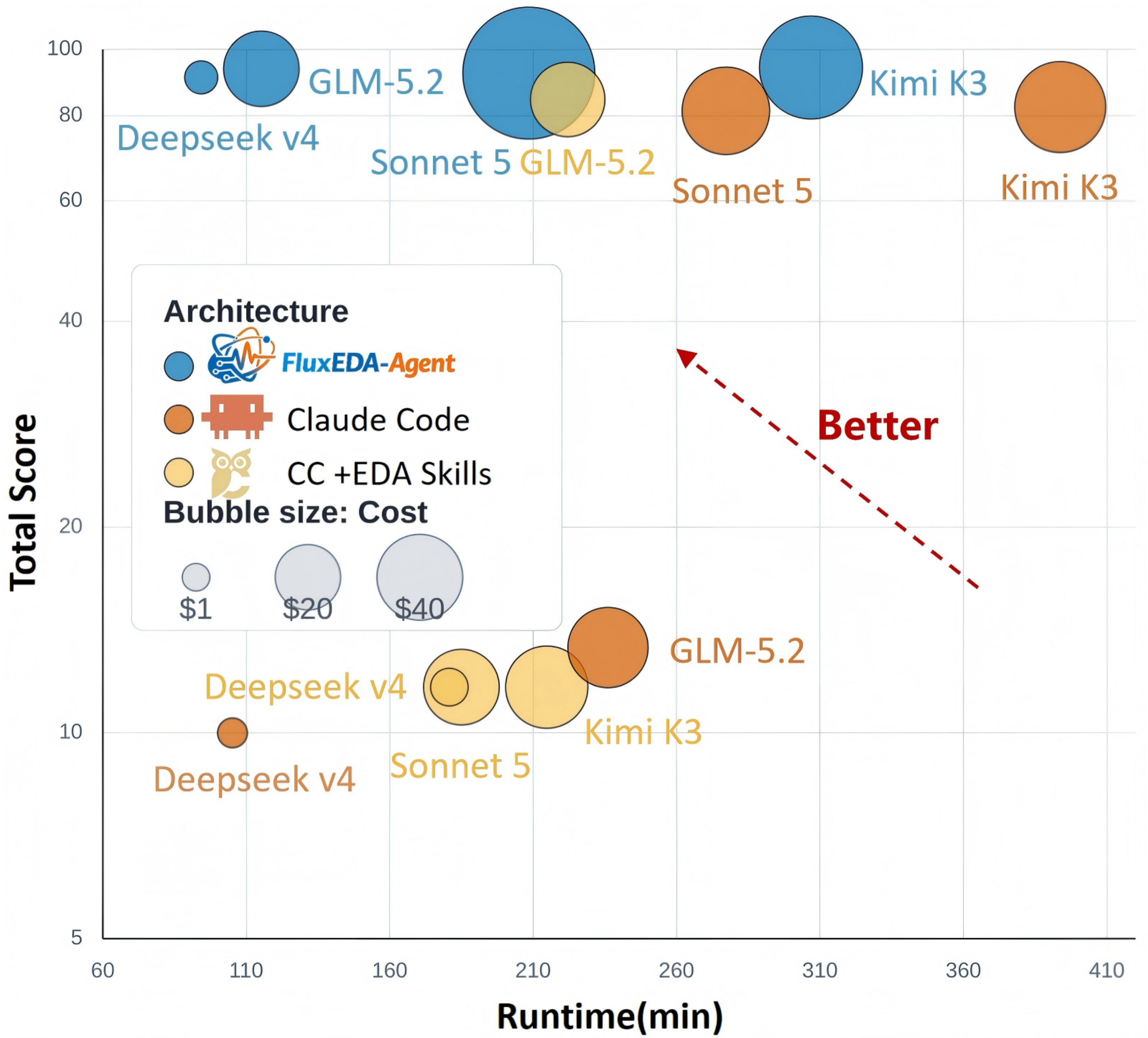}
\caption{Performance--runtime--cost trade-off across agent architectures under the 700 MHz timing target. Higher scores and shorter runtimes are preferred.}
\label{fig:model_and_arch}
\vspace{-0.7cm}
\end{figure}

\begin{table*}[t]
\centering
\caption{End-to-end PicoRV32 RTL-to-GDS results. ROI denotes the normalized Token ROI.}
\label{tab:main-results}
\scriptsize
\setlength{\tabcolsep}{9pt}
\renewcommand{\arraystretch}{1.12}
\resizebox{\textwidth}{!}{%
\begin{tabular}{ll*{8}{c}}
\toprule
\multirow{2}{*}{\textbf{Architecture}}
& \multirow{2}{*}{\textbf{Model}}
& \multicolumn{4}{c}{\textbf{Loose target (350 MHz)}}
& \multicolumn{4}{c}{\textbf{Tight target (700 MHz)}} \\
\cmidrule(lr){3-6}
\cmidrule(lr){7-10}

&
& Score
& Cost (\$)
& Runtime (min)
& ROI
& Score
& Cost (\$)
& Runtime (min)
& ROI \\
\midrule

\multirow{4}{*}{Claude Code}
& DeepSeek-V4
& 13.33 & \textbf{0.11} & \textbf{58.78} & 12.71
& 10.00 & \textbf{0.21} & 105.19 & 5.83 \\

& GLM-5.2
& 13.33 & 33.29 & 302.30 & 0.12
& 13.33 & 14.35 & 236.09 & 0.32 \\

& Claude Sonnet 5
& 87.05 & 20.52 & 284.23 & 1.26
& 81.33 & 18.16 & 277.18 & 1.36 \\

& Kimi K3
& 13.33 & 19.33 & 251.73 & 0.23
& 82.33 & 20.05 & 393.75 & 0.91 \\

\midrule

\multirow{4}{*}{CC+EDA Skills}
& DeepSeek-V4
& 11.67 & 0.64 & 111.70 & 4.81
& 11.67 & 0.94 & 180.81 & 2.75 \\

& GLM-5.2
& 95.97 & 8.37 & 129.84 & 6.22
& 84.47 & 11.56 & 222.04 & 2.61 \\

& Claude Sonnet 5
& 11.67 & 16.96 & 118.81 & 0.42
& 11.67 & 12.11 & 185.00 & 0.40 \\

& Kimi K3
& 13.33 & 29.77 & 472.13 & 0.09
& 11.67 & 15.26 & 214.74 & 0.29 \\

\midrule

\multirow{4}{*}{FluxEDA}
& DeepSeek-V4
& 97.94 & 0.56 & 120.83 & \textbf{39.91}
& 90.99 & 0.45 & \textbf{94.28} & \textbf{48.06} \\

& GLM-5.2
& 97.72 & 10.67 & 110.63 & 5.77
& 93.70 & 11.92 & 115.27 & 4.84 \\

& Claude Sonnet 5
& 96.41 & 48.24 & 252.65 & 0.68
& 92.36 & 51.58 & 208.48 & 0.72 \\

& Kimi K3
& \textbf{98.26} & 18.65 & 176.53 & 2.36
& \textbf{94.10} & 27.81 & 306.83 & 0.95 \\

\bottomrule
\end{tabular}%
}
\end{table*}

\section{Experiments}
\subsection{End-to-End Quality and Cost Efficiency}
Table~\ref{tab:main-results} reports the twelve architecture--model combinations evaluated under two timing targets. Across all four model backends and both targets, every FluxEDA-backed run achieves a score above 90, demonstrating consistently strong end-to-end completion under different timing constraints. This consistency is notable under the tighter 700-MHz constraint,
which places greater pressure on timing closure and physical-design
decisions.

The results also show that cost and runtime cannot be interpreted independently of outcome quality. For example, Claude Code with DeepSeek-V4 is the least expensive and fastest configuration under the 350-MHz target, but achieves a score of only 13.33. Similarly, several expensive runs fail to complete critical stages, resulting in low scores and poor Token ROI. Thus, lower token consumption or shorter runtime does not necessarily indicate greater efficiency when a run fails to produce a valid end-to-end design.

Token ROI provides a more informative measure by relating token expenditure to verified design progress. Configurations that avoid repeated failures, unnecessary retries, and prolonged tool-runtime waste achieve better efficiency than those that consume fewer resources but fail at critical stages. Overall, meaningful cost efficiency depends on reliably converting computational expenditure into validated RTL-to-GDS outcomes.

\subsection{Model Backend Comparison}

\textbf{Kimi K3 achieves the strongest overall performance with FluxEDA, obtaining the highest score under both timing targets.} However, \textbf{without a structured execution architecture, its execution is less stable}. In the Claude Code and CC+EDA Skills configurations, its performance varies substantially across timing targets, with some runs failing to complete required physical-design milestones, such as I/O placement or physical-aware ECO. This contrast indicates that \textbf{FluxEDA makes Kimi K3's high performance reliable through persistent design-state management, stage-aware execution, and physically grounded feedback}.

Among the other backends, DeepSeek-V4 offers the best cost efficiency, GLM-5.2 demonstrates strong cross-architecture adaptability, and Claude Sonnet 5 delivers competitive quality at a higher cost. Overall, strong model reasoning requires structured architectural support to produce reliable end-to-end results.

\subsection{Stage Completion and Workflow Stability}
Table~\ref{tab:stage_completion} summarizes gated stage completion
across the eight model--target runs for each architecture. All runs
complete synthesis, but the command-line-based architectures show
substantially lower completion rates in later stages. Claude Code
completes physical implementation in seven runs and physical-aware
ECO in three, while CC+EDA Skills completes these stages in three
and two runs, respectively. FluxEDA completes all three stages in all eight evaluated runs. All six physical-implementation failures in the command-line-based configurations involve incomplete I/O placement, and one of these runs also fails at CTS. No failures are observed in standard-cell placement or routing, while physical-aware ECO remains the most frequent late-stage failure.

\begin{table}[t]
\centering
\caption{Gated stage completion across eight model--target runs
for each architecture. A stage is counted as complete only when
all required milestones and prerequisite stages are completed.}
\label{tab:stage_completion}
\begin{tabular}{lccc}
\toprule
Architecture & Syn. & Physical Impl. & ECO \\
\midrule
Claude Code     & 8/8 & 7/8 & 3/8 \\
CC + EDA Skills & 8/8 & 3/8 & 2/8 \\
FluxEDA         & 8/8 & 8/8 & 8/8 \\
\bottomrule
\end{tabular}
\vspace{-0.1cm}
\end{table}

Adding EDA skills improves selected configurations, but the gains are
not consistent across models or timing targets. For example,
CC+EDA Skills with GLM-5.2 completes all three stages under both
timing targets, whereas several other skill-augmented configurations
still fail during physical implementation or ECO. These results suggest
that providing procedural EDA knowledge does not by itself ensure
execution reliability. A skill may recommend an appropriate flow or
command family, yet it cannot guarantee that the command is valid for
the active tool version and execution mode, that the required database
state is available, or that the workflow can recover after an error.

Overall, the stage-completion results indicate that reliable
RTL-to-GDS execution depends not only on model knowledge, but also on
the execution support used to preserve design context, enforce valid
stage transitions, and manage tool interactions.

\subsection{Where Do Agents Fail?}
\begin{figure}[t]
\centering
\includegraphics[width=0.9\linewidth]{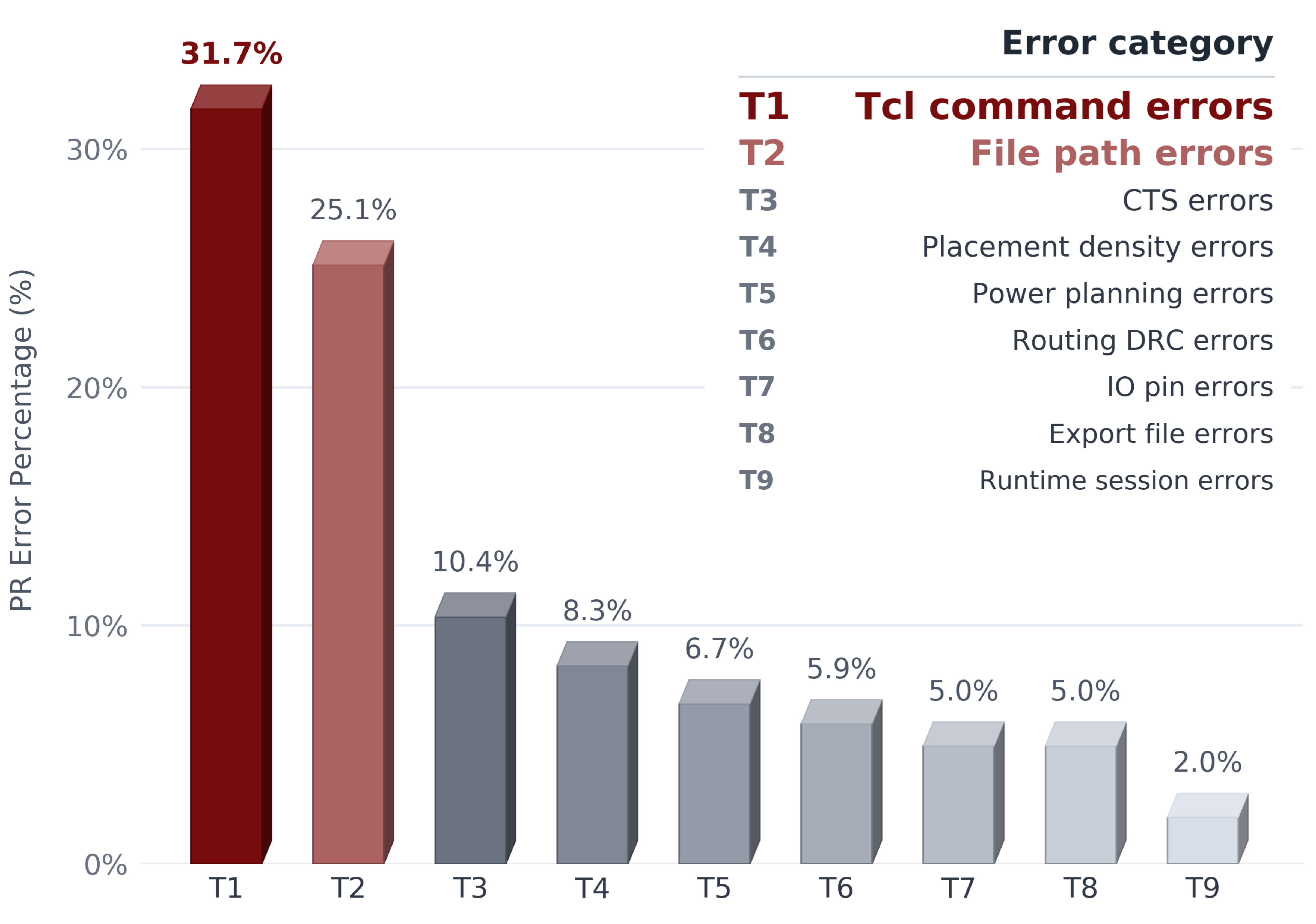}
\caption{Distribution of physical-implementation error categories. The dominant category is tool- and mode-dependent Tcl command compatibility.}
\label{fig:errors}
\vspace{-0.4cm}
\end{figure}

Physical design errors provide a more concrete explanation for the performance gap. As shown in \Cref{fig:errors}, Tcl command compatibility is the largest PR-stage error category, accounting for 31.7\% of observed events. The failures include commands incompatible with the active EDA tool version, execution mode, MMMC initialization, or reporting/export interface. Claude Code with DeepSeek-V4 and GLM-5.2 contributes 48.7\% and 28.0\% of these events, respectively. EDA Skills reduce the fractions to 10.0\% and 7.7\%, but do not eliminate the issue. FluxEDA records 0.0\% for this category across the evaluated models.
The broader lesson is that a general agent often treats the EDA shell as an open programming surface, while the real action space is constrained by tool version, loaded design, flow stage, and project conventions. Structured wrappers and registered APIs can remove entire classes of syntactically plausible but operationally invalid actions.

\section{Lessons for Agentic EDA and Benchmark Design}

\subsection{Domain Skills Do Not Guarantee End-to-End Completion}
The paired architecture comparisons show that the effect of EDA skills is
strongly model- and target-dependent. At 350~MHz, adding EDA skills raises the
GLM-5.2 score from 13.33 to 95.97, but lowers the Claude Sonnet 5 score from
87.05 to 11.67. At 700~MHz, the same intervention raises GLM-5.2 from 13.33 to
84.47, while reducing Kimi K3 from 82.33 to 11.67. The stage profiles show, among the evaluated skill-augmented configurations, only GLM-5.2
completes all three gated stages under both timing targets.

The architecture comparison further indicates that agent capability is a
system-level property rather than a property of the foundation model alone.
For DeepSeek-V4 at 350~MHz, the score spans from 11.67 to 97.94 across the three
architectures, a difference of 86.27 points. Moreover, all eight FluxEDA
model--target runs score above 90 and complete every reported stage. Agentic EDA benchmarks should therefore evaluate complete agent
configurations across multiple models and constraints, rather than
reporting only a single favorable pairing.

\subsection{Efficiency Must Be Conditioned on Verified Progress}
Raw cost and runtime are insufficient measures of end-to-end efficiency
unless interpreted together with verified progress. At 350~MHz, Claude Code
with DeepSeek-V4 is the cheapest and fastest configuration, costing
\$0.11 and taking 58.78 minutes, but it scores only 13.33 and does not
complete physical-aware ECO. FluxEDA with the same model costs \$0.56 and
takes 120.83 minutes, yet completes every reported stage and reaches a score of
97.94 with a normalized Token ROI of 39.91. Early termination can therefore
look inexpensive while producing little end-to-end value.

Token ROI also distinguishes runs that reach the same verified milestone with
very different resource use. At 350~MHz, Claude Code with DeepSeek-V4 and
CC+EDA Skills with Kimi K3 both score 13.33; both complete synthesis and physical implementation, but not physical-aware ECO. Their normalized Token ROI values
are 12.71 and 0.09, respectively, a difference of approximately
$141.2\times$. Conversely, Token ROI should be interpreted only together with stage completion and final design quality: a low-cost partial run may still obtain a
numerically favorable ROI. Benchmarks should therefore jointly report the stage-gated score, milestone
completion, token cost, runtime, and Token ROI.

\subsection{Distinguishing Operational Validity from Optimization Effectiveness}
The error analysis in \Cref{fig:errors} reveals a practical mismatch between
open-ended code generation and state-constrained EDA execution. A Tcl command may be syntactically correct yet
invalid for the active tool version, design state, or flow stage, causing an
agent to fail before its optimization strategy is exercised. Benchmarks should
therefore distinguish operational validity---whether actions are legal and
state-appropriate---from optimization effectiveness---whether valid actions
improve QoR and complete the required stages. A minimal operational contract
covering tool state, allowed actions, required deliverables, and standardized
results makes this distinction auditable without prescribing a particular
backend.

\section{Discussions and Conclusions}
This focused case study supports a clear answer to the title question. Current AI agents \emph{can} complete RTL-to-GDS, but reliable completion is not determined by foundation-model capability alone. The workflow exposes failure modes that are largely invisible in static RTL generation: invalid vendor commands, lost or inconsistent design context, late-stage flow failures, QoR regressions, and inefficient trial-and-error interaction. Domain-specific skills can improve local procedural knowledge, yet they do not by themselves guarantee stable operation across models and timing targets.

The strongest result is therefore not simply the highest numerical score. It is the consistent completion of synthesis, physical implementation, and ECO across model backends and constraints, together with substantially higher Token ROI and the elimination of a dominant class of Tcl compatibility errors. The findings suggest that Agentic EDA should be viewed as a systems problem involving four coupled elements: a capable foundation model, reusable EDA knowledge, structured tool interfaces, and an execution layer that preserves context and controls side effects.

This study is limited to one design, one technology setup, selected commercial-tool versions, and one run per configuration. Although PicoRV32 is more representative than a small test circuit, it does not capture the scale and signoff complexity of industrial SoCs. The results should therefore be interpreted as evidence of workflow behavior rather than universal rankings. Future evaluations should include multiple designs, technologies, toolchains, and repeated runs. They should also cover code generation and repair, multi-stage implementation, and report-based optimization, with controlled ablations of EDA skills, structured APIs, persistent context, checkpoint and rollback mechanisms.

In summary, the central question for AI-assisted chip design is no longer only whether an LLM can generate correct RTL or Tcl. The more consequential question is whether an agent can move a design through a long, heterogeneous toolchain reliably, efficiently, and verifiably. Our results indicate that progress will depend not only on foundation
models, but also on a shift from prompt-centric automation toward
execution-aware infrastructure and process-level benchmarking.

\clearpage
\bibliographystyle{ACM-Reference-Format}
\bibliography{references}
\clearpage

\end{document}